\documentclass{article} % For LaTeX2e
\usepackage{iclr2019_conference,times}

\usepackage{amsmath,amsfonts,bm}

\def\eqref#1{equation~\ref{#1}}
\def\1{\bm{1}}

\DeclareMathAlphabet{\mathsfit}{\encodingdefault}{\sfdefault}{m}{sl}
\SetMathAlphabet{\mathsfit}{bold}{\encodingdefault}{\sfdefault}{bx}{n}

\usepackage[]{hyperref}
\usepackage{url}
\usepackage[title]{appendix}

\usepackage[utf8]{inputenc} % allow utf-8 input
\usepackage[T1]{fontenc}    % use 8-bit T1 fonts
\usepackage{hyperref}       % hyperlinks
\usepackage{url}            % simple URL typesetting
\usepackage{booktabs}       % professional-quality tables
\usepackage{amsmath}
\usepackage{amssymb}
\usepackage{amsfonts}
\usepackage{amstext}
\usepackage{amsthm}
\usepackage{mathrsfs}
\usepackage{nicefrac}       % compact symbols for 1/2, etc.
\usepackage{microtype}      % microtypography
\usepackage{algorithmic}
\usepackage{algorithm}
\usepackage{bm,dsfont}
\usepackage{multirow}
\usepackage{tablefootnote,enumitem}
\usepackage{graphicx}
\usepackage{xcolor}
\usepackage{wrapfig}

\usepackage[capitalise]{cleveref}

\newtheorem{definition}{Definition}

\newtheorem{thm}{Theorem}

\newtheorem{assum}{Assumption}

\newcommand{\xb}{\mathbf{x}}

\newcommand{\RR}{\mathds{R}}

\newcommand{\inner}[2]{\langle #1, #2 \rangle}

\title{Toward Understanding the Impact of \\
Staleness in Distributed Machine Learning}

\author{Wei Dai$^*$, Yi Zhou$^\dagger$, Nanqing Dong$^*$, Hao Zhang$^*$, Eric P. Xing$^*$ \\
$^*$Petuum Inc. \& $^\dagger$Department of Electrical and Computer Engineering, The Ohio State University\\
}

\iclrfinalcopy % Uncomment for camera-ready version, but NOT for submission.
\begin{document}

\maketitle
\vspace{-7pt}
\begin{abstract}
Many distributed machine learning (ML) systems adopt the non-synchronous execution in order to alleviate the network communication bottleneck, resulting in {\it stale} parameters that do not reflect the latest updates. Despite much development in large-scale ML, the effects of staleness on learning are inconclusive as it is challenging to directly monitor or control staleness in complex distributed environments. In this work, we study the convergence behaviors of a wide array of ML models and algorithms under delayed updates. Our extensive experiments reveal the rich diversity of the effects of staleness on the convergence of ML algorithms and offer insights into seemingly contradictory reports in the literature. The empirical findings also inspire a new convergence analysis of stochastic gradient descent in non-convex optimization under staleness, matching the best-known convergence rate of $\mathcal{O}(1/{\sqrt{T}})$.
\end{abstract}
\vspace{-7pt}
\section{Introduction}
\vspace{-7pt}
With the advent of big data and complex models, there is a growing body of works on scaling machine learning under synchronous and non-synchronous\footnote{We use the term ``non-synchronous'' to include both fully asynchronous model~\citep{hogwild} and bounded asynchronous models such as Stale Synchronous Parallel~\citep{ssp}.} distributed execution~\citep{distbelief,1hour,muli_ps}. These works, however, point to seemingly contradictory conclusions on whether non-synchronous execution outperforms synchronous counterparts in terms of absolute convergence, which is measured by the wall clock time to reach the desired model quality. For deep neural networks, \citet{adam,distbelief} show that fully asynchronous systems achieve high scalability and model quality, but others argue that synchronous training converges faster~\citep{revisit,geeps}.
%benefit from lower synchronization overheads, leading to high scalability.
%On the other hand, recent works show that synchronous training of neural networks achieves higher absolute convergence than non-synchronous counterparts, due to the additional computation needed under asynchrony~\citep{revisit,geeps}.
The disagreement goes beyond deep learning models: \citet{ssp,async_ADMM,slow_learners,async_nonconvex,hogwild} empirically and theoretically show that many algorithms scale effectively under non-synchronous settings, but \citet{adarevision,beget_momentum,omnivore} demonstrate significant penalties from asynchrony.

The crux of the disagreement lies in the trade-off between two factors contributing to the absolute convergence: {\it statistical efficiency} and {\it system throughput}. Statistical efficiency measures convergence per algorithmic step (e.g., a mini-batch), while system throughput captures the performance of the underlying implementation and hardware. Non-synchronous execution can improve system throughput due to lower synchronization overheads, which is well understood~\citep{ssp,revisit,cui_atc_14,adam}. However, by allowing various workers to use {\it stale} versions of the model that do not always reflect the latest updates, non-synchronous systems can exhibit lower statistical efficiency~\citep{revisit,geeps}.
%To be sure, there is a broad consensus that lower synchronization results in lower system overhead, which in turn improves system throughput~\citep{ssp,revisit,cui_atc_14,adam}. It is also clear that asynchrony can potentially lead to slower convergence per algorithmic iteration, which may require additional iterations to overcome~\citep{revisit,geeps}. 
%This is a major missing piece in understanding ML convergence in non-synchronous distributed systems: to what extent do stale models impact the statistical efficiency of ML convergence?
How statistical efficiency and system throughput trade off in distributed systems, however, is far from clear.

%The goal is to, by using more machines, improve the absolute convergence speed, as measured by the wall clock time to reach a certain model quality. 

%These contradictory reports can be attributed to a number of factors. 

%Statistical efficiency and distributed system couple

%The difficulty arise because ML convergence couples with the underlying system through staleness. % is due to the entanglement between the ML conver and system throughput via delay, manifested in staleness during convergence.

%The difficulties in understanding the trade-off arise because statistical efficiency and system throughput are coupled in complex distributed environments. 

%It is difficult to understand the impact of staleness in the empirical settings because 

%A key challenge lies in the difficulty of measuring staleness in complex distributed environments.
The difficulties in understanding the trade-off arise because statistical efficiency and system throughput are coupled during execution in distributed environments.
Non-synchronous executions are in general non-deterministic, which can be difficult to profile. Furthermore, large-scale experiments are sensitive to the underlying hardware and software artifacts, which confounds the comparison between studies. Even when they are controlled, innocuous change in the system configurations such as adding more machines or sharing resources with other workloads can inadvertently alter the underlying staleness levels experienced by ML algorithms, masking the true effects of staleness.

%can implicitly change the staleness level experienced by ML algorithms and confound comparison between studies. 

%It is difficult to study the trade-off in distributed systems because it is difficult to control. 

%Staleness is key to disentangle 

%It is difficult to assess the impact of staleness on machine learning (ML) algorithms. First, staleness generally co-occurs with asynchrony in distributed systems, which lead to indeterministic execution that are challenging to profile. Furthermore, the performance of distributed systems is inherently coupled with the hardware environments and software artifacts, making it difficult to compare results across studies%, and therefore good performance on one cluster does not always translate to other environments. 
%Finally, distributed systems are complex. Oftentimes innocuous change in the system configurations such as adding more machines or sharing resources with other workloads can alter the underlying staleness levels. These factors can interact in ways that are difficult to reproduce, hiding the true effects of staleness on ML algorithms.

Understanding the impact of staleness on ML convergence independently from the underlying distributed systems is a crucial step towards decoupling statistical efficiency from the system complexity. 
%is crucial to scale ML algorithms in distributed settings.
%The insights can also serve as target SLA for distributed systems to satisfy.
The gleaned insights can also guide distributed ML system development, potentially using different synchronization for different problems.
%different system designs and optimization for different problems\footnote{For example, the choice of synchronous vs non-synchronous execution can lead to very different system designs and optimization.
%In terms of programming abstraction, non-synchronous systems may restrict client-side updates to be commutative and associative, or requiring additional control on synchronization such as the staleness parameter~\citep{ssp,essp}.
%In terms of system optimization, non-synchronous systems can use early communication to reduce staleness~\citep{bosen}, whereas synchronous system tends to batch updates till clock boundaries.
%}.
In particular, we are interested in the following aspects: Do ML algorithms converge under staleness? To what extent does staleness impact the convergence?

By resorting to simulation study,
%in which staleness level can be controlled and isolated,
we side step the challenges faced in distributed execution.
%in which staleness level can be exactly controlled. These experiments allow us to understand to what extent the staleness impacts the convergence of ML algorithms.
We study the impact of staleness on a diverse set of models: Convolutional Neural Networks (CNNs), Deep Neural Networks (DNNs), multi-class Logistic Regression (MLR), Matrix Factorization (MF), Latent Dirichlet Allocation (LDA), and Variational Autoencoders (VAEs). They are addressed by 7 algorithms, spanning across optimization, sampling, and blackbox variational inference. 
Our findings suggest that while some algorithms are more robust to staleness, no ML method is immune to the negative impact of staleness. We find that all investigated algorithms reach the target model quality under moderate levels of staleness, but the convergence can progress very slowly or fail under high staleness levels. The effects of staleness are also problem dependent. For CNNs and DNNs, the staleness slows down deeper models much more than shallower counterparts. For MLR, a convex objective, staleness has minimal effect. %Empirical gradient coherence analyses 
%For problems with more complex structures like VAE, the convergence slow down due to staleness is much more prominent. 
%Some algorithms are also much more sensitive to staleness than others. 
%Staleness also impacts some algorithms more than others.
Different algorithms respond to staleness very differently.
For example, high staleness levels incur drastically more statistical penalty for RMSProp~\citep{rmsprop} and Adam~\citep{adam_opt} optimization than stochastic gradient descent (SGD) and Adagrad~\citep{adagrad}, which are robust to staleness. %performs well under low staleness, but can be 
%are drastically slowed down by high staleness. Stochastic gradient descent (SGD) and Adagrad~\citep{adagrad}, on the contrary, are overall robust against staleness.
%though its absolute convergence speed is not excellent compared with Adam and other advanced learning rate schedules.
%Outside of optimization,
Separately, Gibbs sampling for LDA is highly resistant to staleness up to a certain level, beyond which it does not converge to a fixed point. 
%and  then undergoes a rapid degradation.
Overall, it appears that staleness is a key governing parameter of ML convergence.

To gain deeper insights, for gradient-based methods we further introduce {\it gradient coherence} along the optimization path, and show that gradient coherence is a possible explanation for an algorithm's sensitivity to staleness. In particular, our theoretical result establishes the $\mathcal{O}(1/\sqrt{T})$ convergence rate of the asynchronous SGD in nonconvex optimization by exploiting gradient coherence, matching the rate of best-known results~\citep{async_nonconvex}.

%For variants of stochastic gradient descent algorithms—now a staple in large-scale ML problems, our results show that even some are much more sensitive to staleness than others.

%We study X algorithm X models. 
% get conclusion paragraph from thesis
%We find that staleness has negative impacts on all studied scenarios. Our study reveals that staleness impacts 

%the impact of staleness is highly problem dependent. 

%To understand the impact of staleness on ML convergence, 
\vspace{-7pt}
\section{Related Work}
\vspace{-7pt}
%There are substantial evidence on both sides of the contention around staleness.
Staleness is reported to help absolute convergence for distributed deep learning in~\citet{adam,distbelief} and has minimal impact on convergence~\citep{beget_momentum,omnivore,async_nonconvex}. But \citet{revisit,geeps} show significant negative effects of staleness. LDA training is generally insensitive to staleness~\citep{yahooLDA10,lightlda,bosen,ssp}, and so is MF training~\citep{nomad,graphlab12,cui_atc_14,async_ADMM}. However, none of their evaluations quantifies the level of staleness in the systems. By explicitly controlling the staleness, we decouple the distributed execution, which is hard to control, from ML convergence outcomes.

We focus on algorithms that are commonly used in large-scale optimization~\citep{1hour,revisit,distbelief}, instead of methods specifically designed to minimize synchronization~\citep{willie,scott2016bayes,jordan2013statistics}. Non-synchronous execution has theoretical underpinning~\citep{li2014communication,ssp,async_ADMM,async_nonconvex,hogwild}. Here we study algorithms that do not necessarily satisfy assumptions in their analyses.

\vspace{-7pt}
\section{Methods}
\vspace{-7pt}
We study six ML models
%that are subject to large-scale application and research~\citep{}. We 
and focus on algorithms that lend itself to data parallelism, which a primary approach for distributed ML. Our algorithms span optimization, sampling, and black box variational inference. %that is a hybrid of optimization and sampling method.in 
\cref{tbl:empirical_scope} summarizes the studied models and algorithms.

\noindent{\bf Simulation Model.} Each update generated by worker $p$ needs to be propagated to both worker $p$'s model cache and other worker's model cache. We apply a uniformly random delay model to these updates that are in transit. Specifically, let $u_p^t$ be the update generated at iteration $t$ by worker $p$. For each worker $p'$ (including $p$ itself), our delay model applies a delay $r_{p,p'}^t\sim \textrm{Categorical}(0,1,..,s-1)$, where $s$ is the maximum delay and $\textrm{Categorical}()$ is the categorical distribution placing equal weights on each integer\footnote{We find that geometrically distributed delays have qualitatively similar impacts on convergence. We defer read-my-write consistency to future work.}. Under this delay model, update $u_p^t$ shall arrive at worker $p'$ at the start of iteration $t+1+r_{p,p'}^t$. The average delay under this model is $\frac{1}{2}s+1$. Notice that for one worker with $s=0$ we reduce to the sequential setting. Since model caches on each worker are symmetric, we use the first worker's model to evaluate the model quality. Finally, we are most interested in measuring convergence against the logical time, and wall clock time is in general immaterial as the simulation on a single machine is not optimized for performance. 

\vspace{-7pt}
\subsection{Models and Algorithms}
\vspace{-7pt}

\begin{table}[ht!]
\centering
\small
\begin{tabular}{ %|c|>{\raggedright}p{3cm}|>{\raggedright}p{4.5cm}|c| }
|c|p{3cm}|p{4.5cm}|c| }
 \hline
 Model & Algorithms & Key Parameters & Dataset \\ \hline \hline
 %MLR & SGD
 %SGD with Momentum\citep{momentum_sgd}, Adam\citep{adam_opt}, AdaDelta\citep{adadelta}, RMSProp\citep{rmsprop}, FTRL\citep{ftrl}
 %& $\eta=0.01$ & MNIST\citep{mnist} \\ \hline
 \multirow{6}{*}{CNN} & SGD & $\eta=0.01$ & \multirow{6}{*}{CIFAR10}\\ \cline{2-3}
 & Momentum SGD & $\eta=0.01$, momentum=0.9 & \\ \cline{2-3}
 & Adam & $\eta=0.001,\beta_1=0.9,\beta_2=0.999$ & \\ \cline{2-3}
 & Adagrad & $\eta=0.01$ & \\ \cline{2-3}
 & RMSProp & $\eta=0.01$, decay=0.9, momentum=0 &\\ \hline
 \multirow{2}{*}{DNN/MLR} & SGD & $\eta=0.01$ & \multirow{2}{*}{MNIST}\\ \cline{2-3}
 & Adam & $\eta=0.001,\beta_1=0.9,\beta_2=0.999$ & \\ \hline
 LDA & Gibbs Sampling & $\alpha=0.1,\beta=0.1$ & 20 NewsGroup \\ \hline
 MF & SGD & $\eta=0.005$, rank=5, $\lambda=0.0001$ & MovieLens1M \\ \hline
%  VAE & VI\tablefootnote{Variational Inference} & Optimization parameters same as MLR/DNN & MNIST\citep{mnist} \\ \hline
 \multirow{2}{*}{VAE} & Blackbox VI (SGD, Adam) & Optimization parameters same as MLR/DNN & \multirow{2}{*}{MNIST} \\ \hline
\end{tabular}
\vspace{-5pt}
\caption{\small Overview of the models, algorithms~\citep{momentum_sgd,adagrad,adam_opt,rmsprop,gibbs_lda}, and dataset~\citep{cifar,mnist,movielens,20news} in our study. $\eta$ denotes learning rate (empirically chosen), $\beta_1,\beta_2$ are optimization hyperparameters (using common default values). $\alpha,\beta$ in LDA are Dirichlet priors for document topic and word topic random variables, respectively.}
\vspace{-10pt}
\label{tbl:empirical_scope}
\end{table}

\noindent{\bf Convolutional Neural Networks (CNNs)} have been a strong focus of large-scale training, both under synchronous~\citep{1hour,geeps,cots} and non-synchronous~\citep{adam,distbelief,revisit,omnivore} training. We consider residual networks with $6n+2$ weight layers~\citep{resnet}. The networks consist of 3 groups of $n$ residual blocks, with 16, 32, and 64 feature maps in each group, respectively, followed by a global pooling layer and a softmax layer.
%The last residual block of each group has stride 2 to reduce spatial resolution.
The residual blocks have the same construction as in~\citep{resnet}. We measure the model quality using test accuracy. For simplicity, we omit data augmentation in our experiments.

\noindent{\bf Deep Neural Networks (DNNs)} are neural networks composed of fully connected layers. %DNNs offer the opportunity to increase the model complexity by adding additional layers, and we can use this tuning knob to study the interactions between model complexity and staleness.
%DNNs are also a type of latent space model, where neurons in the hidden layers are in principle unidentifiable, which can pose challenge to asynchronous execution. 
Our DNNs have 1 to 6 hidden layers, with 256 neurons in each layer, followed by a softmax layer. We use rectified linear units (ReLU) for nonlinearity after each hidden layer~\citep{relu}. {\bf Multi-class Logistic Regression (MLR)} is the special case of DNN with 0 hidden layers. We measure the model quality using test accuracy. %The pseudo-code, similar to MLR, is presented in \cref{alg:mlr}.

\noindent{\bf Matrix factorization (MF)} is commonly used in recommender systems and have been implemented at scale~\citep{nomad,graphlab12,cui_atc_14,async_ADMM,strads_sys,ssp,fugue}. %such as recommending movies to users on Netflix.
Let $D \in \mathbb{R}^{M\times N}$ be a partially filled matrix, MF factorizes $D$ into two factor matrices $L\in \mathbb{R}^{M\times r}$ and $R\in \mathbb{R}^{N\times r}$ %such that their product approximates the ratings: $D \approx LR^T$, where matrix $L\in \mathbb{R}^{M\times r}$ and $R\in \mathbb{R}^{N\times r}$,
($r\ll \min(M,N)$ is the user-defined rank).
% which determines the model size (along with $M$ and $N$).
The $\ell_2$-penalized optimization problem is: $\min_{L,R} \frac{1}{|D_{obs}|} \left\{\sum_{(i,j)\in D_{obs}} ||D_{ij} - \sum_{k=1}^K L_{ik}R_{kj}||^2 + \lambda (||L||_F^2 + ||R||_F^2) \right\}$
where $||\cdot||_F$ is the Frobenius norm and $\lambda$ is the regularization parameter. We partition observations $D$ to workers while treating $L,R$ as shared model parameters. We optimize MF via SGD, and measure model quality by training loss defined by the objective function above. 
%MF is another commonly studied model in the distributed ML literature~\citep{strads_sys,ssp,fugue}. While it is non-convex, it is bi-convex which can have a simpler problem structure than general non-convex problems. %\todo{minibatch sizes}
%We optimize MF with stochastic gradient descent (SGD) 

\noindent{\bf Latent Dirichlet Allocation (LDA)} is an unsupervised method to uncover hidden semantics (``topics'') from a group of documents, each represented as a bag of tokens. In LDA each token $w_{ij}$ ($j$-th token in the $i$-th document) is assigned with a latent topic $z_{ij}$ from totally $K$ topics. We use Gibbs sampling to infer the topic assignments $z_{ij}$. The Gibbs sampling step involves three sets of parameters, known as sufficient statistics: (1) document-topic vector $\theta_i\in \mathbb{R}^K$ where $\theta_{ik}$ the number of topic assignments within document $i$ to topic $k=1...K$; (2) word-topic vector $\phi_w\in \mathbb{R}^K$ where $\phi_{wk}$ is the number of topic assignments to topic $k=1,...,K$ for word (vocabulary) $w$ across all documents; (3) $\tilde{\phi} \in\mathbb{R}^K$ where $\tilde{\phi}_k=\sum_{w=1}^W \phi_{wk}$ is the number of tokens in the corpus assigned to topic $k$. The corpus ($w_{ij}, z_{ij}$) is partitioned to workers,
%(i.e each node has a set of documents), and $\theta_i$ is computed on-the-fly before sampling tokens in document $i$.
while $\phi_w$ and $\tilde{\phi}$ are shared model parameters. We measure the model quality using log likelihood. LDA has been scaled under non-synchronous execution~\citep{yahooLDA12,graphlab12,lightlda} with great success. %To our knowledge there is no report that LDA fails to converge due to staleness.

\noindent{\bf Variational Autoencoder (VAE)} %is an unsupervised model that leverages deep neural networks to construct (non-linear) encoding and decoding functions, in which the encoder function embed objects $\bm{x}$ into a latent code $\bm{z}$~\citep{vae}. VAE can be 
is commonly optimized by black box variational inference, which can be considered as a hybrid of optimization and sampling methods. The inputs to VAE training include two sources of stochasticity: the data sampling $\bm{x}$ and samples of random variable $\epsilon$.
%, which has implication on the gradient behaviors. It represents an interesting hybrid of optimization and sampling method.
We measure the model quality by test loss. We use DNNs with 1$\sim$3 layers as the encoders and decoders in VAE, in which each layer has 256 units furnished with rectified linear function for non-linearity. The model quality is measured by the training objective value, assuming continuous input $\bm{x}$ and isotropic Gaussian prior $p(\bm{z}) \sim \mathcal{N}(0,\bm{I})$.
%, and the optimization is performed over data input and random samples of variables.

%Such an optimal trade-off leads to the following convergence bound for Async-SGD.
%\begin{align*}
%\min_{0\le k\le T} \mathbb{E} \|\nabla F(\xb_{k})\|^2 
%&\le 2\sqrt{L\big(F(\xb_{0}) - \inf_{\xb} F(\xb) \big)}\tfrac{\sigma}{\mu} \sqrt{\tfrac{\log T}{T}}.
%\end{align*}

%\todo{Include gradient coherence-inspired dynamic staleness scheme and experimental results.}

%We also note that by applying the Chebyshev's inequality, \cref{eq: main} further implies that $\|\nabla F(\xb_{k})\| \to 0$ in probability.
\vspace{-7pt}
\section{Experiments}
\vspace{-7pt}
We use batch size 32 for CNNs, DNNs, MLR, and VAEs\footnote{Non-synchronous execution allows us to use small batch sizes, eschewing the potential generalization problem with large batch SGD~\citep{sharp_optima,revisit_small_batch}.}. For MF, we use batch size of 25000 samples, which is 2.5\% of the MovieLens dataset (1M samples).
%Therefore, every 40 batches is 1 full pass over the data.
We study staleness up to $s=50$ on 8 workers, which means model caches can miss updates up to 8.75 data passes.
%($50\times 2.5\% \times 7$). 
For LDA we use $\frac{D}{10 P}$ as the batch size, where $D$ is the number of documents and $P$ is the number of workers. We study staleness up to $s=20$, which means model caches can miss updates up to 2 data passes. We measure time in terms of the amount of work performed, such as the number of batches processed. 
%\todo{put dataset and parameters here vs in table.}
\begin{figure}[ht!]
\centering
\vspace{-10pt}
\includegraphics[width=1\textwidth]{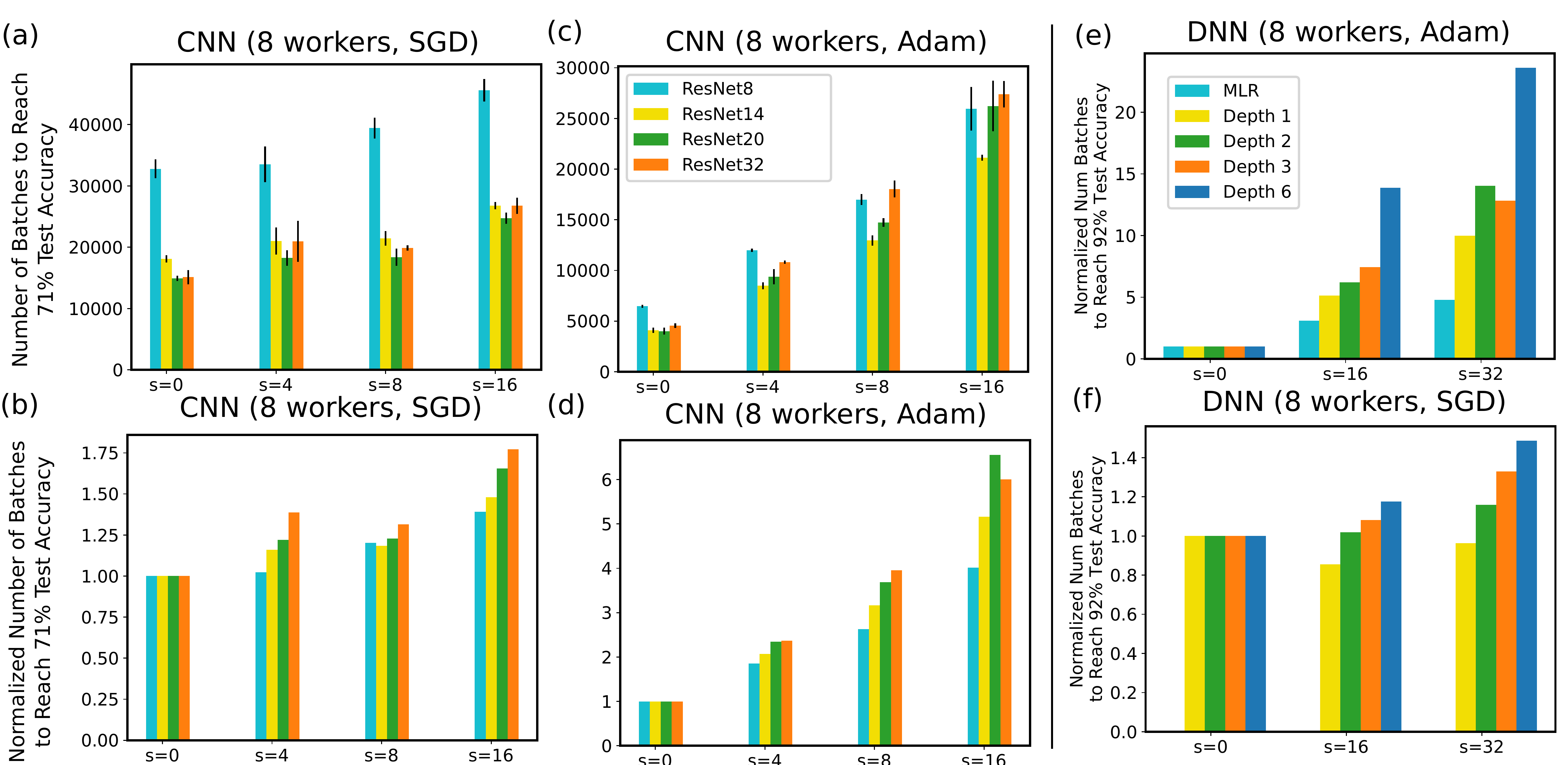}
\vspace{-15pt}
\caption{\small {\bf (a)(c)} The number of batches to reach 71\% test accuracy on CIFAR10 for 4 variants of ResNet with varying staleness, using 8 workers and SGD / Adam. The mean and standard deviation are calculated over 3 randomized runs. {\bf (b)(d)} The same metrics as (a)(c), but each model is normalized by the value under staleness 0 ($s=0$), respectively. {\bf (e)(f)} The number of batches to reach 92\% accuracy for MLR and DNN with varying depths, normalized by the value under staleness 0. MLR with SGD does not converge within the experiment horizon (77824 batches) and thus is omitted in (f).}
\vspace{-10pt}
\label{fig:cnn_depth}
\end{figure}

{\bf Convergence Slowdown.} Perhaps the most prominent effect of staleness on ML algorithms is the slowdown in convergence, evident throughout the experiments. \cref{fig:cnn_depth} shows the number of batches needed to reach the desired model quality for CNNs and DNNs/MLR with varying network depths and different staleness ($s=0,...,16$).
%By examining the normalized plots
\cref{fig:cnn_depth}(b)(d) show that convergence under higher level of staleness requires more batches to be processed in order to reach the same model quality. This additional work can potentially be quite substantial, such as in \cref{fig:cnn_depth}(d) where it takes up to 6x more batches compared with settings without staleness ($s=0$). It is also worth pointing out that while there can be a substantial slowdown in convergence, the optimization still reaches desirable models under most cases in our experiments. %There are notable exceptions where staleness actually accelerates convergence as discussed in the sequel.
When staleness is geometrically distributed (\cref{fig:grad_corr}(c)), we observe similar patterns of convergence slowdown.

We are not aware of any prior work reporting slowdown as high as observed here. This finding has important ramifications for distributed ML. Usually, the moderate amount of workload increases due to parallelization errors can be compensated by the additional computation resources and higher system throughput in the distributed execution. However, it may be difficult to justify spending large amount of resources for a distributed implementation if the statistical penalty is too high, which should be avoided (e.g., by staleness minimization system designs or synchronous execution). %The high level of statistical penalty 
%Thus, proper understanding of the impact of staleness is key to effective large-scale ML.

{\bf Model Complexity.} \cref{fig:cnn_depth} also reveals that the impact of staleness can depend on ML parameters, such as the depths of the networks.
%ResNet and DNNs with varying depths, 
Overall we observe that staleness impacts deeper networks more than shallower ones. This holds true for SGD, Adam, Momentum, RMSProp, Adagrad (\cref{fig:algorithm_time}), and other optimization schemes, and generalizes to other numbers of workers (see Appendix).
%other optimization schemes besides Adam and SGD, including Momentum, Adagrad, RMSProp, and 

This is perhaps not surprising, given the fact that deeper models pose more optimization challenges even under the sequential settings~\citep{glorot_init,resnet}, though we point out that existing literature does not explicitly consider model complexity as a factor in distributed ML~\citep{async_nonconvex,1hour}. Our results suggest that the staleness level acceptable in distributed training can depend strongly on the complexity of the model. For sufficiently complex models it may be more advantageous to eliminate staleness altogether and use synchronous training.

%\todo{Define cosine similarity and connect it}

{\bf Algorithms' Sensitivity to Staleness.} Staleness has uneven impacts on different SGD variants. \cref{fig:algorithm_time} shows the amount of work (measured in the number of batches) to reach the desired model quality for five SGD variants. \cref{fig:algorithm_time}(d)(e)(f) reveals that while staleness generally increases the number of batches needed to reach the target test accuracy, the increase can be drastic for certain algorithms, such as Adam, Momentum, and RMSProp. RMSProp, in particular, fails to converge to the desired test accuracy under several settings. On the other hand, SGD and Adagrad appear to be robust to the staleness, with the total workload <2x of non-stale case ($s=0$) even under high staleness ($s=16$)\footnote{Note that $s=0$ execution treats each worker's update as separate updates instead of one large batch in other synchronous systems.}. In fact, in some cases, the convergence is accelerated by staleness, such as Adagrad on 1 worker under $s=16$. This may be attributed to the implicit momentum created by staleness~\citep{beget_momentum} and the aggressive learning rate shrinking schedule in Adagrad. 

Our finding is consistent with the fact that, to our knowledge, all existing successful cases applying non-synchronous training to deep neural networks use SGD~\citep{distbelief,adam}. In contrast, works reporting subpar performance from non-synchronous training all use variants SGD, such as RMSProp with momentum~\citep{revisit} and momentum~\citep{geeps}. Our results suggest that these different outcomes may be primarily driven by the choice of optimization algorithms, leading to the seemingly contradictory reports of whether non-synchronous execution is advantageous over synchronous ones.

%In the case of FTRL staleness appears to lower the number of batches necessary, potentially due to the implicit momentum created by staleness~\citep{beget_momentum}, and momentum is helpful for convergence as evident in~\cref{fig:measure_stale}(a).

%For example, Adam, SGD with Momentum, RMSProp, and AMSGrad are highly sensitive to the staleness.

{\bf Effects of More Workers.} The impact of staleness is amplified by the number of workers. In the case of MF, \cref{fig:mf_lda_vae}(b) shows that %staleness has a larger impact on 8 workers than 4 workers, as can be seen in the growth of the number of batches to convergence in~\cref{fig:hypo8_norm}. In fact, 
the convergence slowdown in terms of the number of batches (normalized by the convergence for $s=0$) on 8 workers is more than twice of the slowdown on 4 workers. For example, in \cref{fig:mf_lda_vae}(b) the slowdown at $s=15$ is $\sim$3.4, but the slowdown at the same staleness level on 8 workers is $\sim$8.2. Similar observations can be made for CNNs (\cref{fig:mf_lda_vae}). This can be explained by the fact that additional workers amplifies the effect of staleness by (1) generating updates that will be subject to delays, and (2) missing updates from other workers that are subject to delays. %In the sequel we will see how the number of workers amplify the effective staleness in theoretical analyses.

{\bf LDA.} \cref{fig:mf_lda_vae}(c)(d) show the convergence curves of LDA with different staleness levels for two settings varying on the number of workers and topics. Unlike the convergence curves for SGD-based algorithms (see Appendix), the convergence curves of Gibbs sampling are highly smooth, even under high staleness and a large number of workers. This can be attributed to the structure of log likelihood objective function~\citep{gibbs_lda}. Since in each sampling step we only update the count statistics based on a portion of the corpus, the objective value will generally change smoothly.

Staleness levels under a certain threshold ($s\le 10$) lead to convergence, following indistinguishable log likelihood trajectories, regardless of the number of topics ($K=10,100$) or the number of workers (2--16 workers, see Appendix). Also, there is very minimal variance in those trajectories. However, for staleness beyond a certain level ($s\ge 15$), Gibbs sampling does not converge to a fixed point. The convergence trajectories are distinct and are sensitive to the number of topics and the number of workers. There appears to be a ``phase transition'' at a certain staleness level that creates two distinct phases of convergence behaviors\footnote{We leave the investigation into this distinct phenomenon as future work.}. We believe this is the first report of a staleness-induced failure case for LDA Gibbs sampling.

{\bf VAE} In \cref{fig:mf_lda_vae}(e)(f), VAEs exhibit a much higher sensitivity to staleness compared with DNNs (\cref{fig:cnn_depth}(e)(f)). This is the case even considering that VAE with depth 3 has 6 weight layers, which has a comparable number of model parameters and network architecture to DNNs with 6 layers. We hypothesize that this is caused by the additional source of stochasticity from the sampling procedure, in addition to the data sampling process.
%This is consistent with our findings (not shown) that VAEs have much lower gradient coherence than DNNs with comparable setup.
% For example, when optimized by the Adam optimizer, VAEs with depth 3 under $s=8$ requires 17.5x more batches to converge than under $s=0$. On the other hand, DNNs with 6 layers optimized by the Adam optimizer takes only $<$5x more batches to converge under $s=16$ than under $s=0$ (\cref{fig:hypo2_norm}). This high sensitivity of Adam on VAE is a robust phenomenon, since the error bars in the unnormalized plots (\cref{fig:hypo4_unnorm_panel} in the appendix).

%\todo{talk about gradient coherence? Need experiments for 1 worker staleness 4}

\iffalse
\begin{figure}[ht!]
\centering
%\begin{minipage}[l]{0.8\textwidth}
%\centering
\includegraphics[width=0.8\textwidth]{grad_corr.pdf}
%\end{minipage}
%\begin{minipage}[l]{0.19\textwidth}
\caption{\small Cosine similarity between the gradient at the $k$-th iteration $\nabla F(\xb_k)$, and the gradient $m$ steps prior $\nabla F(\xb_{k-m})$. The x-axis is $m=1,...,10$. For computational efficiency, we approximate the full gradient $\nabla F(\xb_k)$ by gradients on a fixed set of 1000 training samples $D_{fixed}$ and use $\nabla_{D_{fixed}} F(\xb_k)$.}
\label{fig:grad_corr}
\vspace{-10pt}
%\end{minipage}
\end{figure}
\fi

\begin{figure}[ht!]
%\vspace{-15pt}
\centering
\includegraphics[width=1\textwidth]{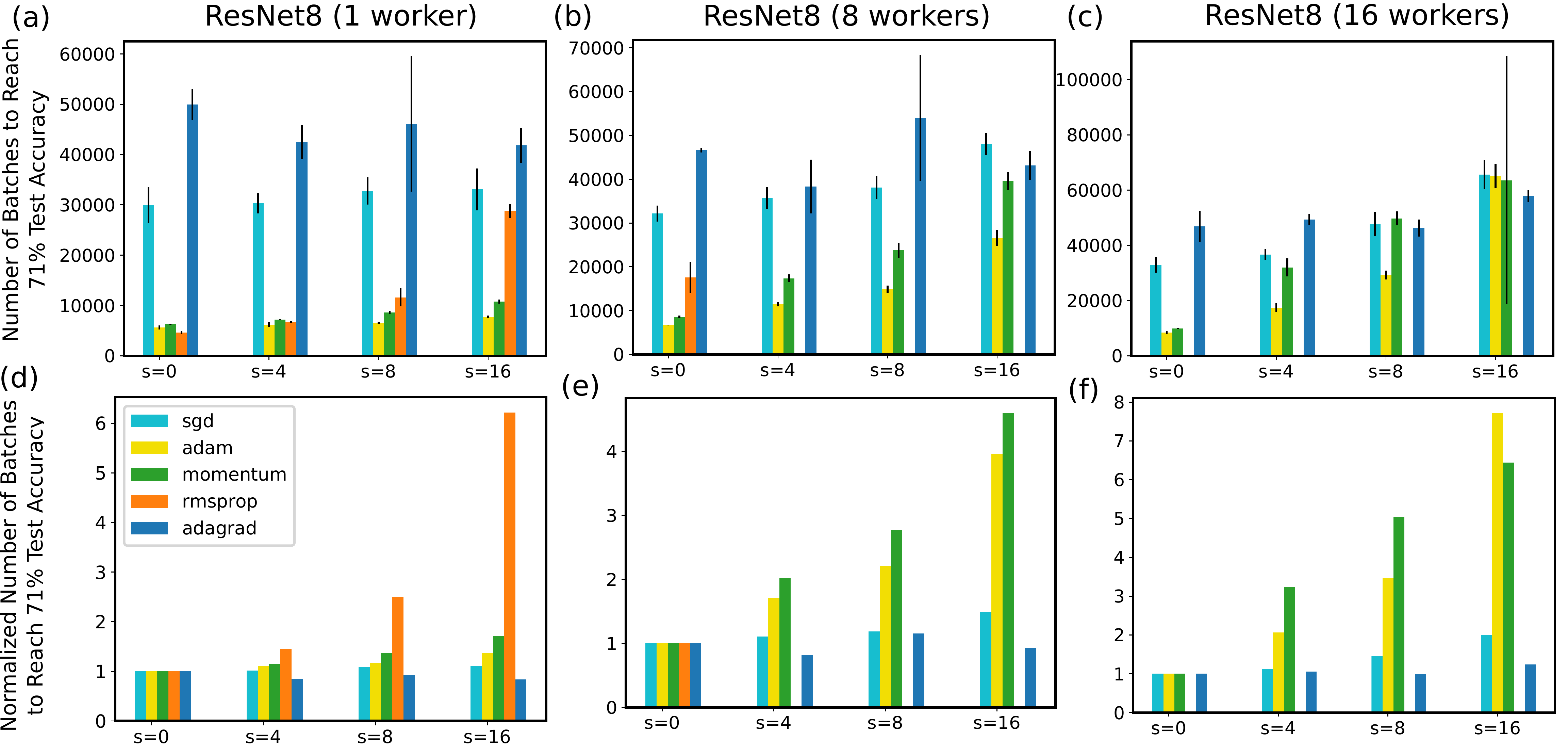}
\vspace{-15pt}
\caption{\small (a)(b)(c) The number of batches to reach 71\% test accuracy on 1, 8, 16 workers with staleness $s=0,...,16$ using ResNet8. We consider 5 variants of SGD: SGD, Adam, Momentum, RMSProp, and Adagrad. The error bars are 1 standard deviation based on 3 independent runs. (d)(e)(f) show the same metric but each algorithm is normalized by the value under staleness 0 ($s=0$), respectively. Under certain settings RMSProp does not converge to the desired model quality within the experiment horizon and is thus omitted.}
\vspace{-10pt}
\label{fig:algorithm_time}
\end{figure}

\begin{figure}[ht!]
\centering
\includegraphics[width=1\textwidth]{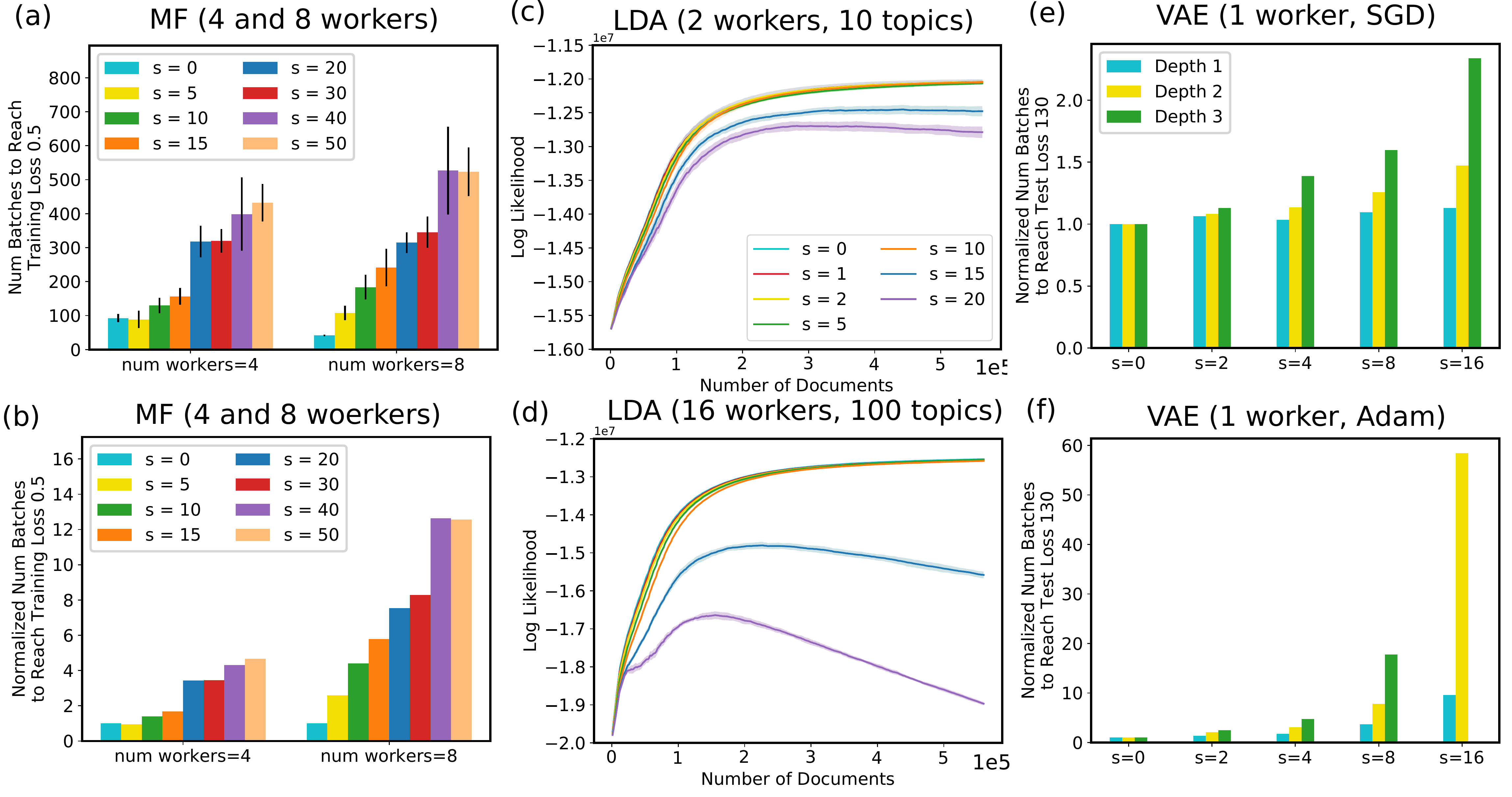}
\vspace{-24pt}
\caption{\small {\bf (a)} The number of batches to reach training loss of 0.5 for Matrix Factorization (MF). {\bf (b)} shows the same metric in (a) but normalized by the values of staleness 0 of each worker setting, respectively (4 and 8 workers). {\bf (c)(d)} Convergence of LDA log likelihood using 10 and 100 topics under staleness levels $s=0,...,20$, with 2 and 16 workers. The convergence is recorded against the number of documents processed by Gibbs sampling. The shaded regions are 1 standard deviation around the means (solid lines) based on 5 randomized runs. (e)(f) The number of batches to reach test loss 130 by Variational Autoencoders (VAEs) on 1 worker, under staleness $s=0,...,16$. We consider VAEs with depth 1, 2, and 3 (the number of layers in the encoder and the decoder networks, separately). The numbers of batches are normalized by $s=0$ for each VAE depth, respectively. Configurations that do not converge to the desired test loss are omitted in the graph, such as Adam optimization for VAE with depth 3 and $s=16$.
%(See Supplemental Materials for the unnormalized version.)
}
\vspace{-15pt}
\label{fig:mf_lda_vae}
\end{figure}

\vspace{-5pt}
\section{Gradient Coherence and Convergence of Asynchronous SGD}
\vspace{-7pt}

We now provide theoretical insight into the effect of staleness on the observed convergence slowdown. We focus on the challenging asynchronous SGD (Async-SGD) case, which characterizes the neural network models, among others.
%In this section, we investigate the factors that affect the convergence of the asynchronous SGD (Async-SGD) from a theoretical perspective.
Consider the following nonconvex optimization problem
%\vspace{-5pt}
\begin{align}
	\min_{\xb \in \RR^d} ~ F(\xb) := \frac{1}{n} \sum_{i=1}^{n} f_i(\xb), \tag{P}
\end{align}
%\vspace{-5pt}
where $f_i$ corresponds to the loss on the $i$-th data sample, and the objective function is assumed to satisfy the following standard conditions:
\begin{assum}\label{assum: F}
	The objective function $F$ in the problem (P) satisfies:
	\begin{enumerate}[leftmargin=*, noitemsep, topsep=0pt]
		\item Function $F$ is continuously differentiable and bounded below, i.e., $\inf_{\xb\in \RR^d} F(\xb) > -\infty$;
		\item The gradient of $F$ is $L$-Lipschitz continuous.
	\end{enumerate}
\end{assum}
Notice that we allow $F$ to be nonconvex. We apply the Async-SGD to solve the problem (P). Let $\xi(k)$ be the mini-batch of data indices sampled from $\{1, \ldots, n\}$ uniformly at random by the algorithm at iteration $k$, and $|\xi(k)|$ is the mini-batch size. Denote mini-batch gradient as $\nabla f_{\xi(k)} (\xb_{k}) := \sum_{i\in \xi(k)}\nabla f_{i} (\xb_{k})$. Then, the update rule of Async-SGD can be written as
\begin{align}
&\xb_{k+1} = \xb_{k} - \tfrac{\eta_k }{|\xi(\tau_k)|} \nabla f_{\xi(\tau_k)} (\xb_{\tau_k}),
%\text{where}~ &k-s+1 \le \tau_k \le k,
\tag{Async-SGD}
\end{align}
where $\eta_k$ corresponds to the stepsize, $\tau_k$ denotes the delayed clock and the maximum staleness is assumed to be bounded by $s$. This implies that $k-s+1 \le \tau_k \le k$. 
%We note that the mini-batch of gradient used in Async-SGD can be understood as the aggregated delayed stochastic gradients on a particular machine in an asynchronous distributed system.  
%Async-SGD agregates the gradients generated at time $\tau_k$ but is applied at time $k$.

The optimization dynamics of Async-SGD is complex due to the nonconvexity and the uncertainty of the delayed updates. 
Interestingly, we find that the following notion of gradient coherence provides insights toward understanding the convergence property of Async-SGD. 
\begin{definition}[Gradient coherence]\label{def:coherence}
	The gradient coherence at iteration $k$ is defined as
	\begin{align*}
		\mu_k := \min_{k-s+1\le t\le k} \tfrac{\inner{\nabla F(\xb_{k})}{\nabla F(\xb_{t})}}{\|\nabla F(\xb_{k})\|^2}.
	\end{align*}
\end{definition}
\vspace{-10pt}
Parameter $\mu_k$ captures the minimum coherence between the current gradient $\nabla F(\xb_{k})$ and the gradients along the past $s$ iterations.  Intuitively, if $\mu_k$ is positive, then the direction of the current gradient is well aligned to those of the past gradients. In this case, the convergence property induced by using delayed stochastic gradients is close to that induced by using synchronous stochastic gradients.
Note that $\mu_k$ is easy to compute empirically during the course of asynchronous optimization\footnote{It can be approximated by storing a pre-selected batch of data on a worker. The worker just needs to compute gradient every $T$ mini-batches to obtain approximate $\nabla F(\xb_k)$, $\nabla F(\xb_t)$ in \cref{def:coherence}.}. Empirically we observe that $\mu_k > 0$ through most of the optimization path, especially when the staleness is minimized in practice by system optimization  (\cref{fig:grad_corr}). Our theory can be readily adapted to account for a limited amount of negative $\mu_k$ (see Appendix), but our primary interest is to provide a quantity that is (1) easy to compute empirically during the course of optimization, and (2) informative for the impact of staleness and can potentially be used to control synchronization levels. We now characterize the convergence property of Async-SGD.

\begin{figure}[ht!]
\centering
%\begin{minipage}[l]{0.8\textwidth}
%\centering
\includegraphics[width=1\textwidth]{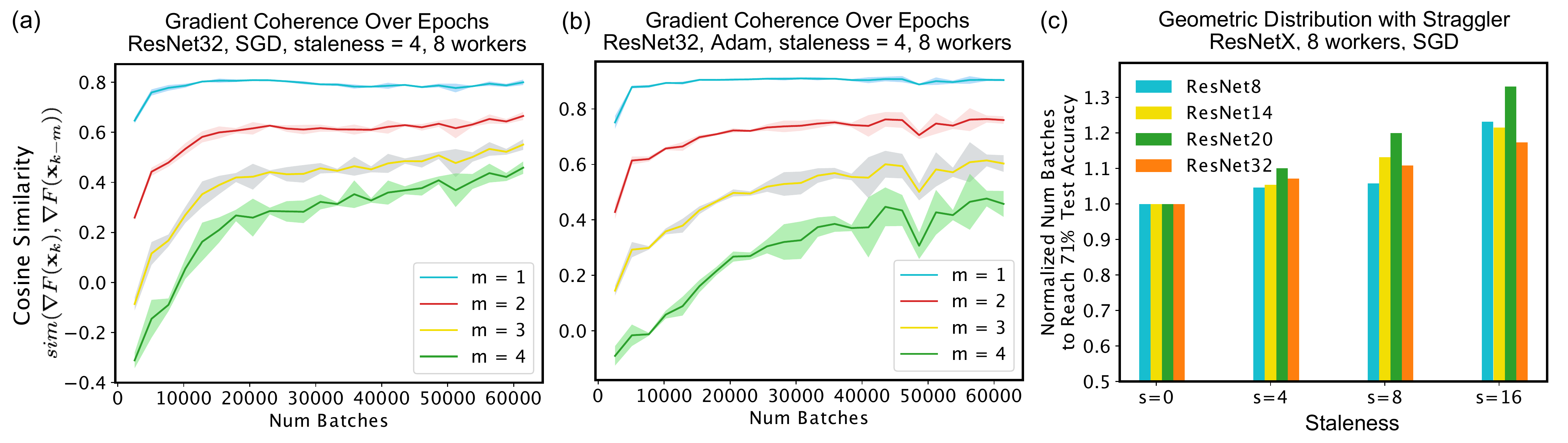}
%\end{minipage}
\vspace{-23pt}
%\begin{minipage}[l]{0.19\textwidth}
\caption{\small (a)(b) Cosine similarity between the gradient at the $k$-th iteration $\nabla F(\xb_k)$, and the gradient $m$ steps prior $\nabla F(\xb_{k-m})$, over the course of convergence for ResNet32 on CIFAR10 optimized by SGD (a) and Adam (b) under staleness $s=4$ on 8 workers with parameters in Table 1. Shaded region is 1 standard deviation over 3 runs. For computational efficiency, we approximate the full gradient $\nabla F(\xb_k)$ by gradients on a fixed set of 1000 training samples $D_{fixed}$ and use $\nabla_{D_{fixed}} F(\xb_k)$. (c) The number of batches to reach 71\% test accuracy on CIFAR10 for ResNet8-32 using 8 workers and SGD under geometric delay distribution (details in Appendix).}
\label{fig:grad_corr}
\vspace{-10pt}
%\end{minipage}
\end{figure}

\begin{thm}\label{thm: main}
	Let \Cref{assum: F} hold. Suppose for some $\mu > 0$, the gradient coherence satisfies $\mu_k\ge \mu$ for all $k$ and the variance of the stochastic gradients is bounded by $\sigma^2 > 0$. Choose stepsize $\eta_k = \frac{\mu}{sL\sqrt{k}}$. Then, the iterates generated by the Async-SGD satisfy
	\begin{align}
		\min_{0\le k\le T} \mathbb{E} \|\nabla F(\xb_{k})\|^2 \le \big(\tfrac{sL(F(\xb_{0}) - \inf_{\xb} F(\xb) )}{\mu^2} + \tfrac{\sigma^2\log T}{s}\big) \tfrac{1}{\sqrt{T}}. \label{eq: main}
	\end{align}
\end{thm}
\vspace{-5pt}
We refer readers to Appendix for the the proof. \Cref{thm: main} characterizes several theoretical aspects of Async-SGD. First, the choice of the stepsize $\eta_k = \frac{\mu}{sL\sqrt{k}}$ is adapted to both the maximum staleness and the gradient coherence. Intuitively, if the system encounters a larger staleness, then a smaller stepsize should be used to compensate the negative effect. On the other hand, the stepsize can be accordingly enlarged if the gradient coherence along the iterates turns out to be high. In this case, the direction of the gradient barely changes along the past several iterations, and a more aggressive stepsize can be adopted. In summary, the choice of stepsize should trade-off between the effects caused by both the staleness and the gradient coherence.

\begin{wrapfigure}{r}{0.5\textwidth}
\vspace{-20pt}
  \begin{center}
    \includegraphics[width=0.48\textwidth]{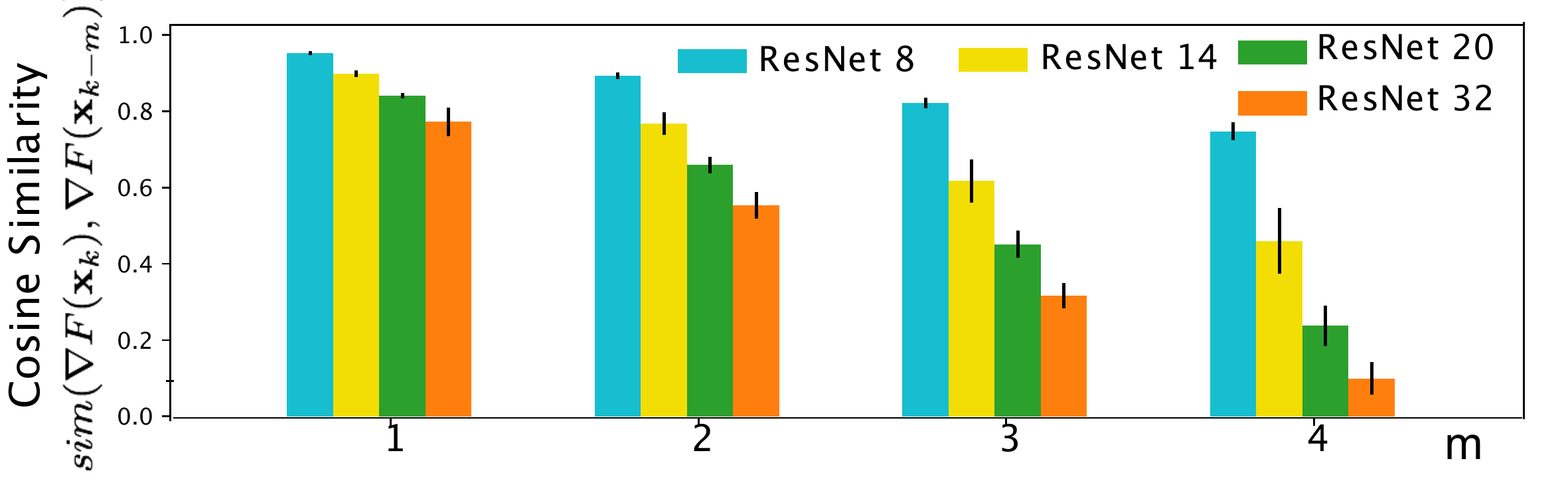}
  \end{center}
  \vspace{-20pt}
  \caption{\small Gradient coherence for ResNet with varying depths optimized by SGD using 8 workers. The x-axis $m$ is defined in \cref{fig:grad_corr}}
  \label{fig:grad_snapshot}
  \vspace{-10pt}
\end{wrapfigure}

Furthermore, \Cref{thm: main} shows that the minimum gradient norm decays at the rate $\mathcal{O}(\frac{\log T}{\sqrt{T}})$, implying that the Async-SGD converges to a stationary point provided a positive gradient coherence, which we observe empirically in the sequel. On the other hand, the bound in \cref{eq: main} captures the trade-off between the maximum staleness $s$ and the gradient coherence $\mu$. Specifically, minimizing the right hand side of \cref{eq: main} with regard to the maximum staleness $s$ yields the optimal choice $s^* = \sigma \mu \sqrt{\frac{\log T}{L(F(\xb_{0}) - \inf_{\xb} F(\xb))}}$, i.e., a larger staleness is allowed if the gradients remain to be highly coherent along the past iterates.

{\bf Empirical Observations.} \cref{thm: main} suggests that more coherent gradients along the optimization paths can be advantageous under non-synchronous execution. \cref{fig:grad_corr} shows the cosine similarity $sim(\bm{a},\bm{b}):= \frac{\bm{a}\cdot\bm{b}}{\|\bm{a}\| \|\bm{b}\|}$ between gradients along the convergence path for CNNs and DNNs\footnote{Cosine similarity is closely related to the coherence measure in \cref{def:coherence}.}. We observe the followings: (1) Cosine similarity improves over the course of convergence (\cref{fig:grad_corr}(a)(b)). Except the highest staleness during the early phase of convergence, cosine similarity remains positive. In practice the staleness experienced during run time can be limited to small staleness~\citep{essp}, which minimizes the likelihood of negative gradient coherence during the early phase. (2) \cref{fig:grad_snapshot} shows that Cosine similarity decreases with increasing model complexity for CNN models. This is consistent with the convergence difficulty encountered in deeper models (\cref{fig:cnn_depth}).

%(1) Except the deepest architectures, cosine similarity remains positive through the staleness window (i.e., $m=1,...,s$ in \cref{fig:grad_corr}), and hence a positive $\mu_k$ in \cref{def:coherence}, supporting the assumption of \cref{thm: main}. (2) Cosine similarity decreases with increasing model complexity for both CNNs and DNNs and can become negatively correlated. This is consistent with the convergence difficulty encountered in deeper models (\cref{fig:cnn_depth}). %\todo{comment on post-convergence gradient coherence?}

\vspace{-7pt}
\section{Discussion and Conclusion}
\vspace{-7pt}

In this work, we study the convergence behaviors under delayed updates for a wide array of models and algorithms. Our extensive experiments reveal that staleness appears to be a key governing parameter in learning. Overall staleness slows down the convergence, and under high staleness levels the convergence can progress very slowly or fail. The effects of staleness are highly problem dependent, influenced by model complexity, choice of the algorithms, the number of workers, and the model itself, among others. Our empirical findings inspire new analyses of non-convex optimization under asynchrony based on gradient coherence, matching the existing rate of $\mathcal{O}(1/\sqrt{T})$.

Our findings have clear implications for distributed ML. To achieve actual speed-up in absolute convergence, any distributed ML system needs to overcome the slowdown from staleness, and carefully trade off between system throughput gains and statistical penalties. Many ML methods indeed demonstrate certain robustness against low staleness, which should offer opportunities for system optimization. Our results support the broader observation that existing successful non-synchronous systems generally keep staleness low and use algorithms efficient under staleness~\citep{muli_ps,ssp}.
%Furthermore, it is paramount to implement the system in a way that minimizes the staleness

%, such as using a more suitable communication protocol or network bandwidth management. 

\bibliography{nips2018}
\bibliographystyle{iclr2019_conference}

\newpage

%\begin{appendices}
\appendix
\section{Appendix}
\subsection{Proof of Theorem 1}

\begin{thm}%\label{thm: main}
	Let Assumption 1 hold. Suppose the gradient coherence $\mu_k$ is lower bounded by some $\mu > 0$ for all $k$ and the variance of the stochastic gradients is upper bounded by some $\sigma^2 > 0$. Choose stepsize $\eta_k = \frac{\mu}{sL\sqrt{k}}$. Then, the iterates generated by the Async-SGD satisfy
	\begin{align}
		\min_{0\le k\le T} \mathbb{E} \|\nabla F(\xb_{k})\|^2 \le \big(\tfrac{sL(F(\xb_{0}) - \inf_{\xb} F(\xb) )}{\mu^2} + \tfrac{\sigma^2\log T}{s}\big) \tfrac{1}{\sqrt{T}}.
		%\label{eq: main}
	\end{align}
\end{thm}
\label{sec: thm: main}
\begin{proof}
	By the $L$-Lipschitz property of $\nabla F$, we obtain that for all $k$
	\begin{align}
		F(\xb_{k+1}) &\le F(\xb_{k}) + \inner{\xb_{k+1} - \xb_{k}}{\nabla F(\xb_{k})} + \frac{L}{2} \|\xb_{k+1} - \xb_{k}\|^2 \\
		&= F(\xb_{k}) - \eta_k \inner{\nabla f_{\xi(\tau_k)} (\xb_{\tau_k})}{\nabla F(\xb_{k})} + \frac{L\eta_k^2}{2} \|\nabla f_{\xi(\tau_k)} (\xb_{\tau_k})\|^2.
	\end{align}
	Taking expectation on both sides of the above inequality and note that the variance of the stochastic gradient is bounded by $\sigma^2$,  we further obtain that
	\begin{align}
	\mathbb{E}	[F(\xb_{k+1})] &\le \mathbb{E} [F(\xb_{k})] - \eta_k \mathbb{E} [\inner{\nabla F (\xb_{\tau_k})}{\nabla F(\xb_{k})}] + \frac{L\eta_k^2}{2} \mathbb{E} [\|\nabla F (\xb_{\tau_k})\|^2 + \sigma^2] \\
	&\le \mathbb{E} [F(\xb_{k})] - \eta_k \mu_k \mathbb{E} \|\nabla F(\xb_{k})\|^2 + \frac{L\eta_k^2}{2} \mathbb{E} \|\nabla F (\xb_{\tau_k})\|^2 + \frac{\sigma^2L\eta_k^2}{2} \\
	&\le \mathbb{E} [F(\xb_{k})] - \eta_k \mu \mathbb{E} \|\nabla F(\xb_{k})\|^2 + \frac{L\eta_k^2}{2} \sum_{t=k-s+1}^{k}\mathbb{E} \|\nabla F (\xb_{t})\|^2 + \frac{\sigma^2L\eta_k^2}{2}.
	\end{align}
Telescoping the above inequality over $k$ from $0$ to $T$ yields that
\begin{align}
	&\mathbb{E}	[F(\xb_{k+1})] - \mathbb{E}	[F(\xb_{0})] \nonumber\\
	&\le - \sum_{k=0}^{T} \eta_k \mu \mathbb{E} \|\nabla F(\xb_{k})\|^2 + \frac{L}{2} \sum_{k=0}^{T}\sum_{t=k-s+1}^{k} \eta_k^2\mathbb{E} \|\nabla F (\xb_{t})\|^2 + \frac{\sigma^2L}{2} \sum_{k=0}^T \eta_k^2 \\
	&\le - \sum_{k=0}^{T} \eta_k \mu \mathbb{E} \|\nabla F(\xb_{k})\|^2 + \frac{L}{2} \sum_{k=0}^{T}\sum_{t=k-s+1}^{k} \eta_t^2\mathbb{E} \|\nabla F (\xb_{t})\|^2 + \frac{\sigma^2L}{2} \sum_{k=0}^T \eta_k^2 \\
	&\le - \sum_{k=0}^{T} \eta_k \mu \mathbb{E} \|\nabla F(\xb_{k})\|^2 + \frac{Ls}{2} \sum_{k=0}^{T} \eta_k^2\mathbb{E} \|\nabla F (\xb_{k})\|^2 + \frac{\sigma^2L}{2} \sum_{k=0}^T \eta_k^2 \\
	&= \sum_{k=0}^{T} \big(\frac{Ls\eta_k^2}{2} - \eta_k \mu\big) \mathbb{E} \|\nabla F(\xb_{k})\|^2 + \frac{\sigma^2L}{2} \sum_{k=0}^T \eta_k^2.
\end{align}	
Rearranging the above inequality and note that $F(\xb_{k+1}) > \inf_{\xb} F(\xb) > -\infty$, we further obtain that
\begin{align}
	\sum_{k=0}^{T} \big(\eta_k \mu -\frac{Ls\eta_k^2}{2} \big) \mathbb{E} \|\nabla F(\xb_{k})\|^2 \le \big(F(\xb_{0}) - \inf_{\xb} F(\xb)\big) + \frac{\sigma^2L}{2} \sum_{k=0}^T \eta_k^2.
\end{align}
Note that the choice of stepsize guarantees that $\eta_k \mu -\frac{Ls\eta_k^2}{2} > 0$ for all $k$. Thus, we conclude that
\begin{align}
	\min_{0\le k\le T} \mathbb{E} \|\nabla F(\xb_{k})\|^2 &\le \frac{\big(F(\xb_{0}) - \inf_{\xb} F(\xb)\big) + \frac{\sigma^2L}{2} \sum_{k=0}^T \eta_k^2}{\sum_{k=0}^{T} \big(\eta_k \mu -\frac{Ls\eta_k^2}{2} \big)} \\
	&\le \frac{2\big(F(\xb_{0}) - \inf_{\xb} F(\xb)\big) + \sigma^2L \sum_{k=0}^T \eta_k^2}{\sum_{k=0}^{T} \eta_k \mu},
\end{align}
where the last inequality uses the fact that $\eta_k \mu -\frac{Ls\eta_k^2}{2} > \frac{\eta_k \mu}{2}$. Substituting the stepsize $\eta_k = \frac{\mu}{sL\sqrt{k}}$ into the above inequality and simplifying, we finally obtain that
\begin{align}
\min_{0\le k\le T} \mathbb{E} \|\nabla F(\xb_{k})\|^2 
&\le \bigg(\frac{sL\big(F(\xb_{0}) - \inf_{\xb} F(\xb) \big)}{\mu^2} + \frac{\sigma^2\log T}{s}\bigg) \frac{1}{\sqrt{T}}.
\end{align}
	
\end{proof}

\subsection{Handling Negative Gradient Coherence in Theorem 1}

Our assumption of positive gradient coherence (GC) is motivated by strong empirical evidence that GC is largely positive (Fig. 4(a)(b) in the main text). Contrary to conventional wisdom, GC generally {\em improves} when approaching convergence for both SGD and Adam. Furthermore, in practice, the effective staleness for any given iteration generally concentrates in low staleness for the non-stragglers~\citep{essp}. 
%Thus, if the effective staleness distribution is concentrated around a small number, the gradient coherence is high on average. We will provide more empirical results on this in the revision. 
%Our assumption is motivated by empirical observations that $\mu_k$ is always positive when the training is close to convergence. Whereas during the training, we observe that a larger staleness leads to $\mu_k$ that is more likely to be negative. We will provide more empirical results to illustrate the coherence in the revision. 
%When $\mu_k$ is negative at some iterations, the key eq. 11 in supplementary is related to .

When some $\mu_k$ are negative at some iterations, in eq. 11 in the Appendix we can move the negative terms in $\sum_k \eta_k \mu_k$ to the right hand side and yield a higher upper bound (i.e., slower convergence). This is also consistent with empirical observations that higher staleness lowers GC and slows convergence.

\subsection{Exponential delay distribution.} We consider delays drawn from geometric distribution (GD), which is the discrete version of exponential distribution. For each iterate we randomly select a worker to be the straggler with large mean delay ($p=0.1$), while all other non-straggler workers have small delays. The non-straggler delay is drawn from GD with $p$ chosen to achieve the same mean delay as in the uniform case (after factoring in straggler) in the main text. The delay is drawn per worker for each iteration, and thus a straggler's outgoing updates to all workers suffer the same delay. 
%For example, $p=0.5$ for non-straggler and $p=0.1$ for straggler achieve the same mean delay as the uniform delay with $s=4$ under 8 workers.
Fig. 4(c) in the main text shows the convergence speed under the corresponding staleness $s$ with the same mean delay (though $s$ is not a parameter in GD). It exhibits trends analogous to Fig. 1(b) in main text: staleness slows convergence substantially and overall impacts deeper networks more.

\subsection{Additional Results for DNNs}

\begin{figure}[t!]
\centering
\includegraphics[width=0.9\textwidth]{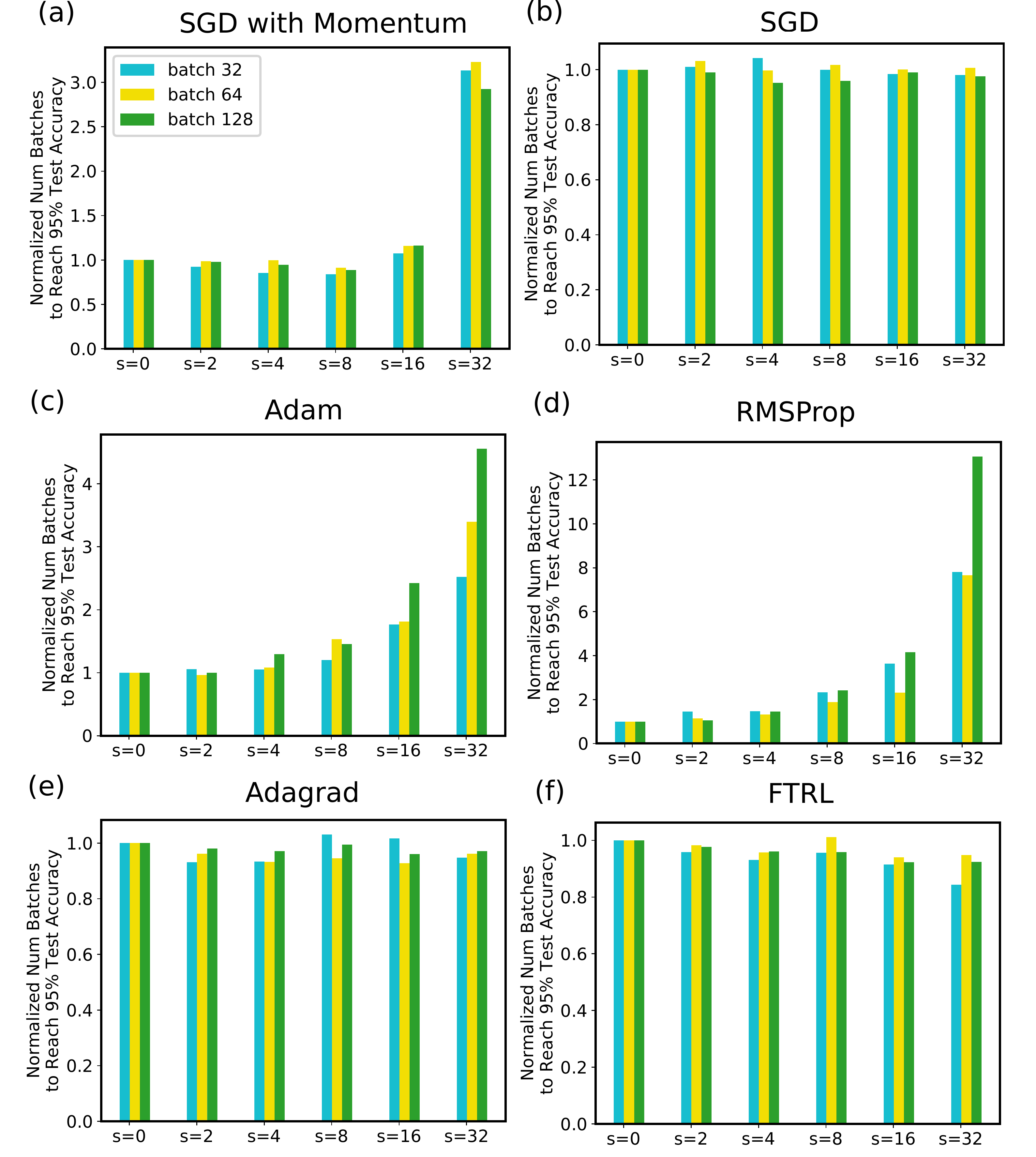}
\caption{The number of batches to reach 95\% test accuracy using 1 hidden layer and 1 worker, respectively normalized by $s=0$.}
\label{fig:measure_stale_norm}
\end{figure}

We present additional results for DNNs. \cref{fig:measure_stale_norm} shows the number of batches, normalized by $s=0$, to reach convergence using 1 hidden layer and 1 worker under varying staleness levels and batch sizes. Overall, the effect of batch size is relatively small except in high staleness regime ($s=32$).

\begin{figure}[ht!]
\centering
\includegraphics[width=0.9\textwidth]{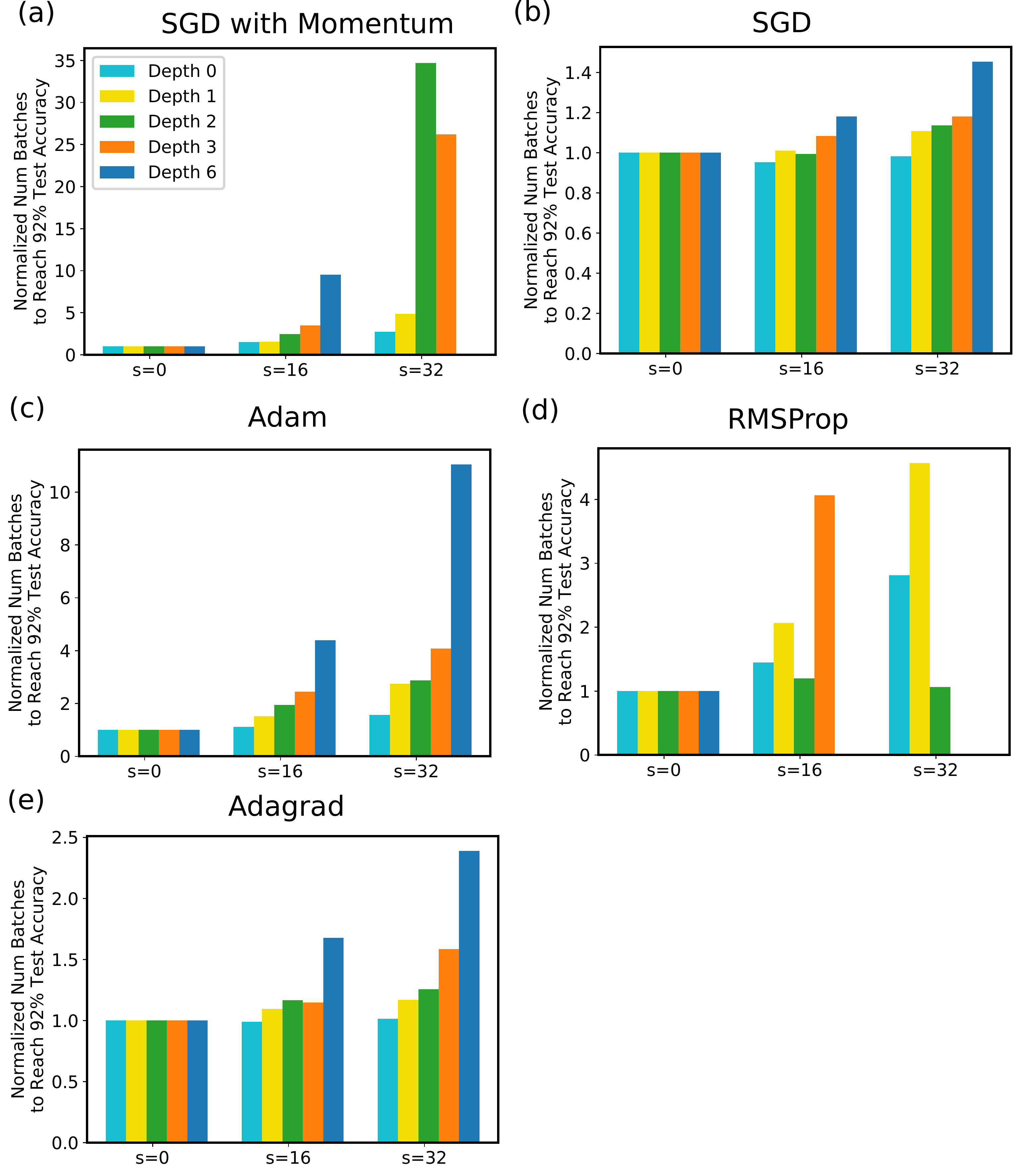}
\caption{The number of batches to reach 92\% test accuracy using DNNs with varying numbers of hidden layers under 1 worker. We consider several variants of SGD algorithms (a)-(e). Note that with depth 0 the model reduces to MLR, which is convex. The numbers are averaged over 5 randomized runs. We omit the result whenever convergence is not achieved within the experiment horizon (77824 batches), such as SGD with momentum at depth 6 and $s=32$. %We do not include FTRL result due to the unstable convergence. The unnormalized version is provided in the appendix (\cref{fig:hypo2}).
}
\label{fig:hypo2_norm}
\end{figure}

\cref{fig:hypo2_norm} shows the number of batches to reach convergence, normalized by $s=0$ case, for 5 variants of SGD using 1 worker. The results are in line with the analyses in the main text: staleness generally leads to larger slow down for deeper networks than shallower ones. SGD and Adagrad are more robust to staleness than Adam, RMSProp, and SGD with momentum. In particular, RMSProp exhibit high variance in batches to convergence (not shown in the normalized plot) and thus does not exhibit consistent trend.

\begin{figure}[ht!]
\centering
\includegraphics[width=0.9\textwidth]{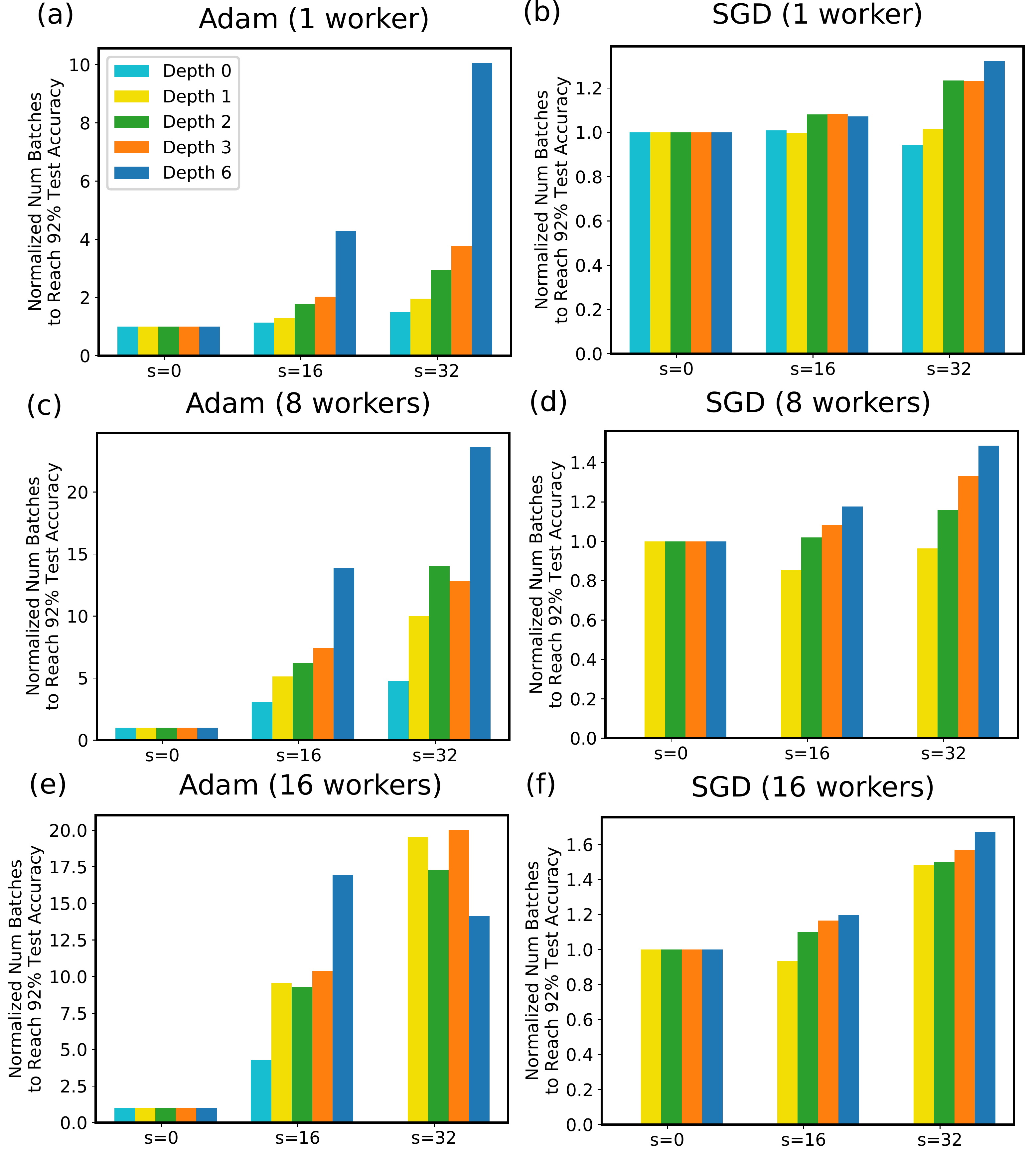}
\caption{The number of batches to reach 92\% test accuracy with Adam and SGD on 1, 8, 16 workers with varying staleness. Each model depth is normalized by the staleness 0's values, respectively. The numbers are average over 5 randomized runs. Depth 0 under SGD with 8 and 16 workers did not converge to target test accuracy within the experiment horizon (77824 batches) for all staleness values, and is thus not shown.
%The unnormalized version is in Appendix (\cref{fig:hypo3_unnorm_panel})
}
\label{fig:hypo3_panel}
\end{figure}

\cref{fig:hypo3_panel} shows the number of batches to convergence under Adam and SGD on 1, 8, 16 simulated workers, respectively normalized by staleness 0's values. The results are consistent with the observations and analyses in the main text, namely, that having more workers amplifies the effect of staleness. We can also observe that SGDS is more robust to staleness than Adam, and shallower networks are less impacted by staleness. In particular, note that staleness sometimes accelerates convergence, such as in \cref{fig:hypo3_panel}(d). This is due to the implicit momentum created by staleness~\citep{beget_momentum}.

%\newpage

\subsection{Additional Results for LDA}

\begin{figure}[ht!]
\centering
\includegraphics[width=0.8\textwidth]{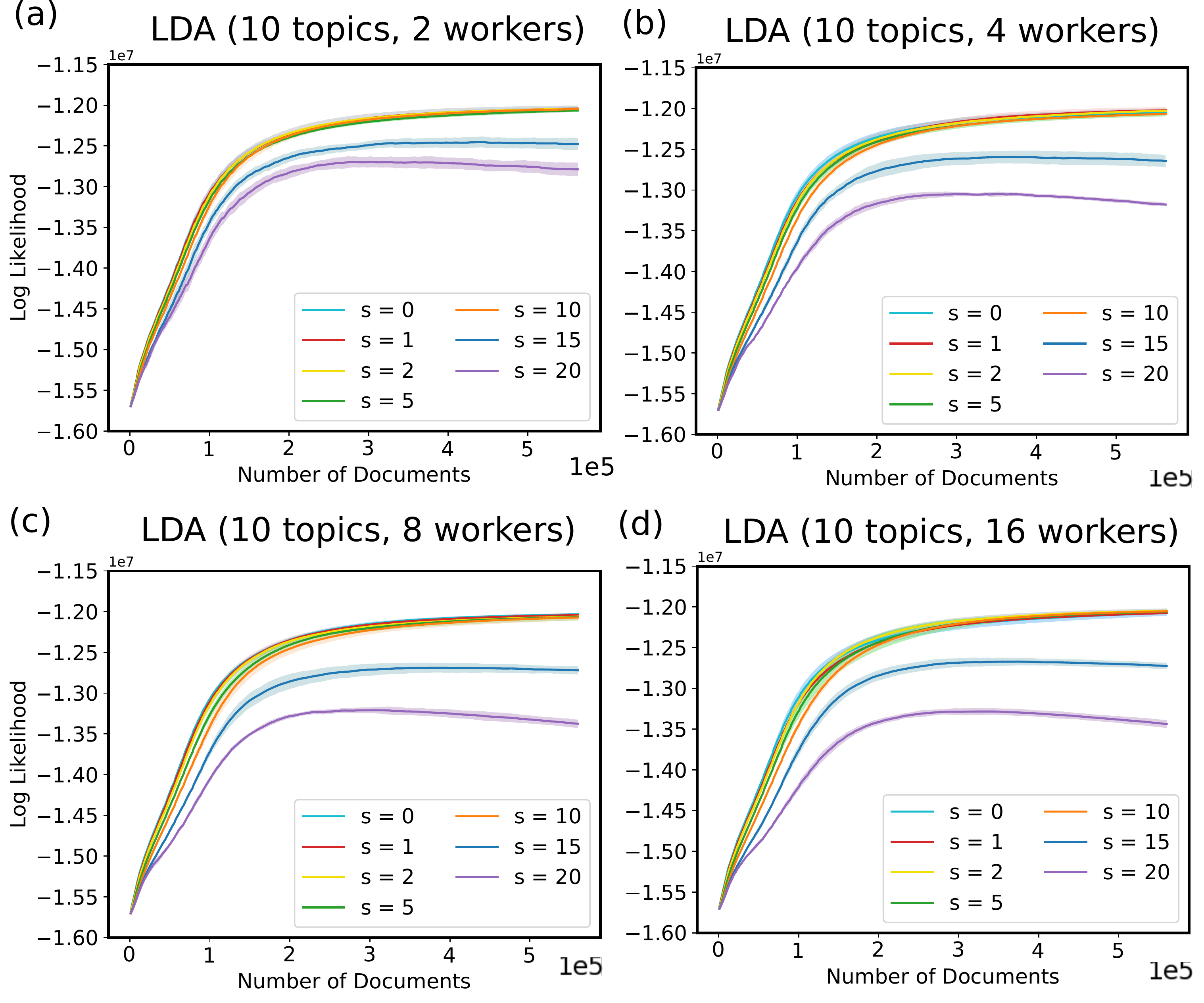}
\caption{Convergence of LDA log likelihood using 10 topics with respect to the number of documents processed by collapsed Gibbs sampling, with varying staleness levels and number of workers. The shaded regions are 1 standard deviation around the means (solid lines) based on 5 randomized runs.}
\label{fig:lda_K10}
\end{figure}

\begin{figure}[ht!]
\centering
\includegraphics[width=0.8\textwidth]{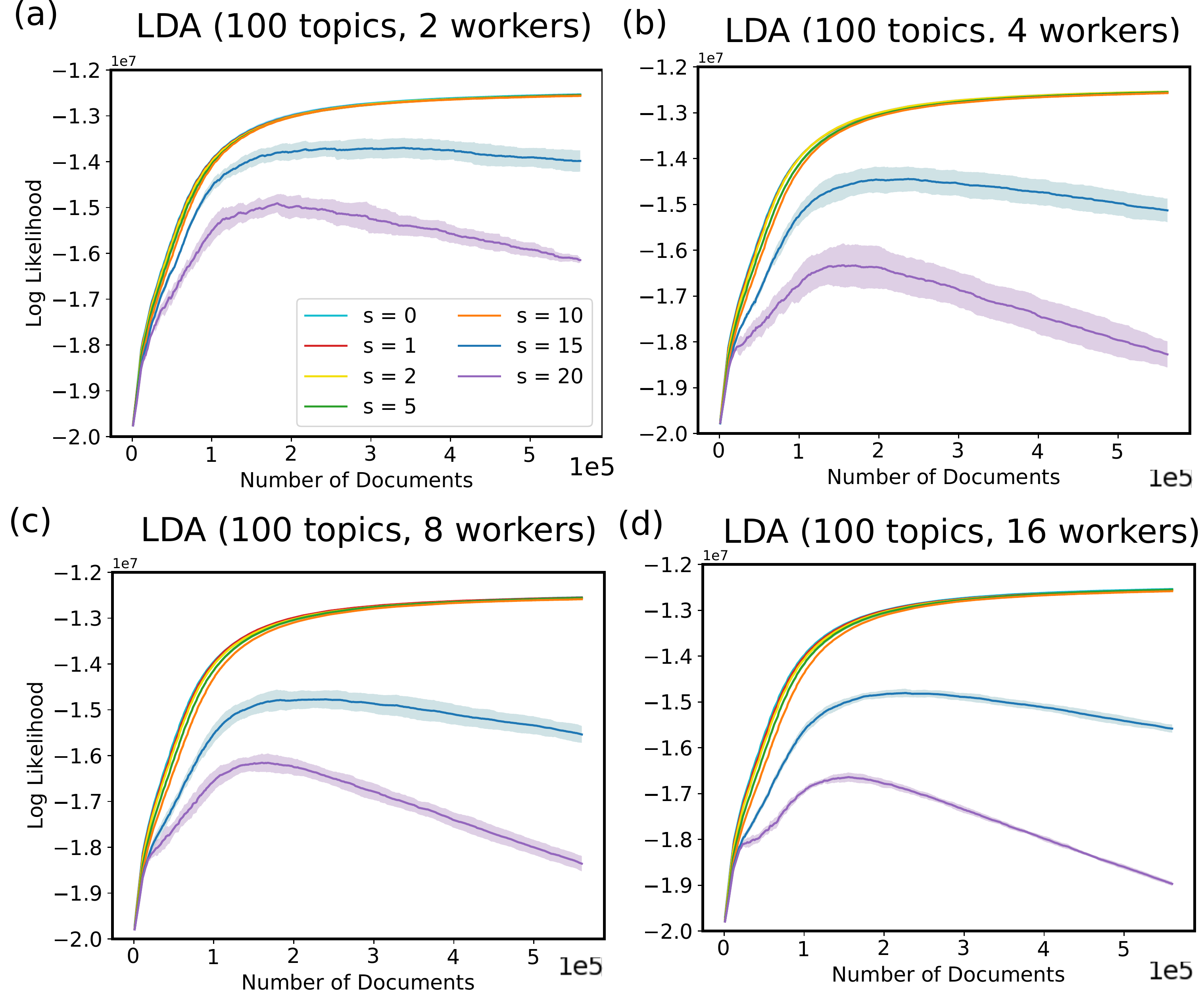}
\caption{Convergence of LDA log likelihood using 100 topics with respect to the number of documents processed by collapsed Gibbs sampling, with varying staleness levels and the number of workers. The shaded regions are 1 standard deviation around the means (solid lines) based on 5 randomized runs.}
\label{fig:lda_K100}
\end{figure}

We present additional results of LDA under different numbers of workers and topics in \cref{fig:lda_K10} and \cref{fig:lda_K100}. These panels extends~\cref{fig:mf_lda_vae}(c)(d) in the main text. See the main text for experimental setup and analyses and experimental setup.

%\newpage

\subsection{Additional Results for MF}

\begin{figure}[ht!]
\centering
\includegraphics[width=0.8\textwidth]{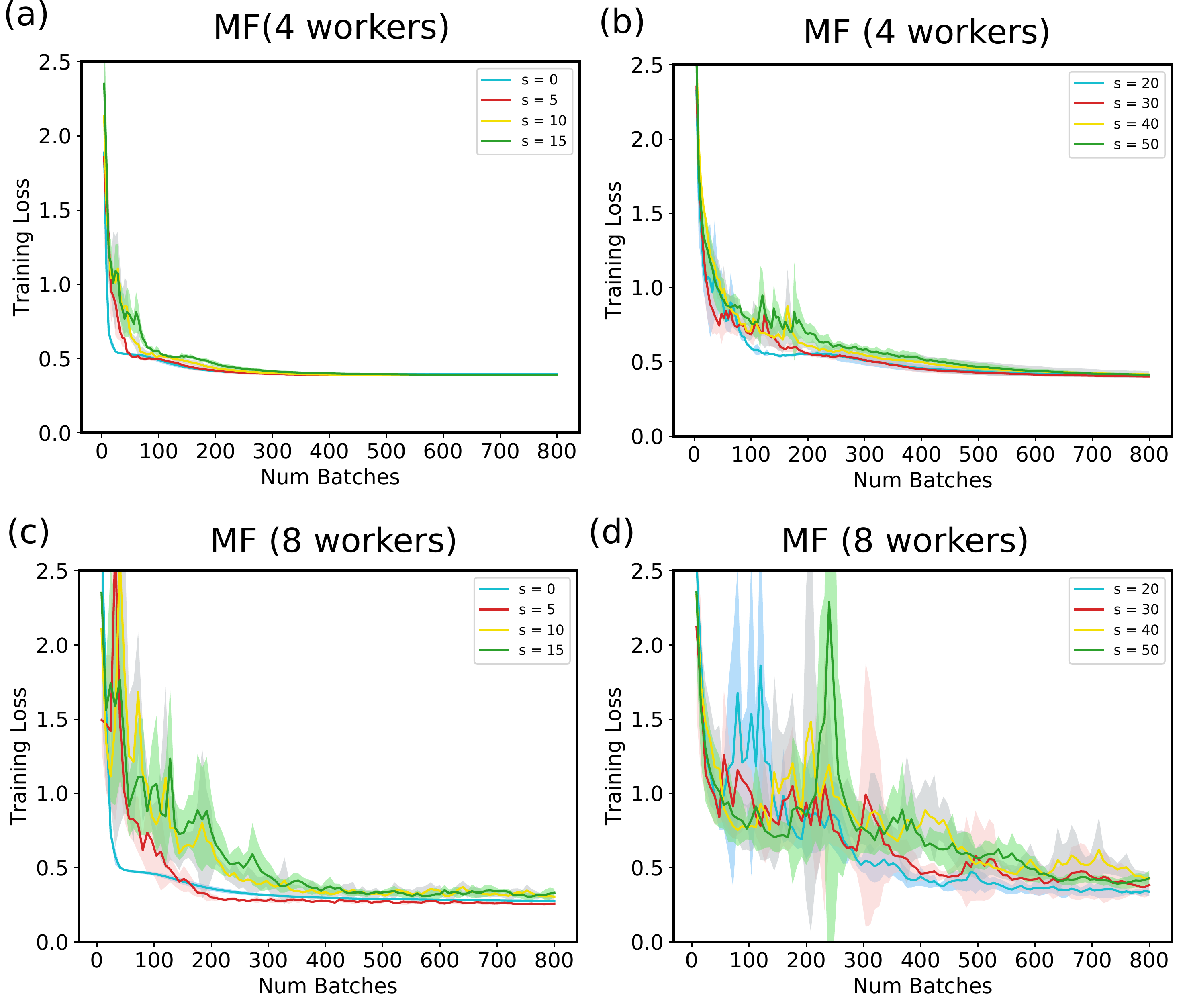}
\caption{Convergence of Matrix Factorization (MF) using 4 and 8 workers, with staleness ranging from 0 to 50.
%We use the number of batches processed across all workers as the logical time.
The x-axis shows the number of batches processed across all workers. Shaded area represents 1 standard deviation around the means (solid curves) computed on 5 randomized runs.}
\label{fig:hypo8_convergence}
\end{figure}

We show the convergence curves for MF under different numbers of workers and staleness levels in \cref{fig:hypo8_convergence}. It is evident that higher staleness leads to a higher variance in convergence.
%This is reflected in the convergence curves (\cref{fig:hypo8_convergence}).
Furthermore, the number of workers also affects variance, given the same staleness level. For example, MF with 4 workers incurs very low standard deviation up to staleness 20. In contrast, MF with 8 workers already exhibits a large variance at staleness 15. The amplification of staleness from increasing number of workers is consistent with the discussion in the main text. See the main text for experimental setup and analyses. %See the main text for the analyses.

%\newpage

\subsection{Additional Results for VAEs}

\begin{figure}[ht!]
\centering
\includegraphics[width=0.9\textwidth]{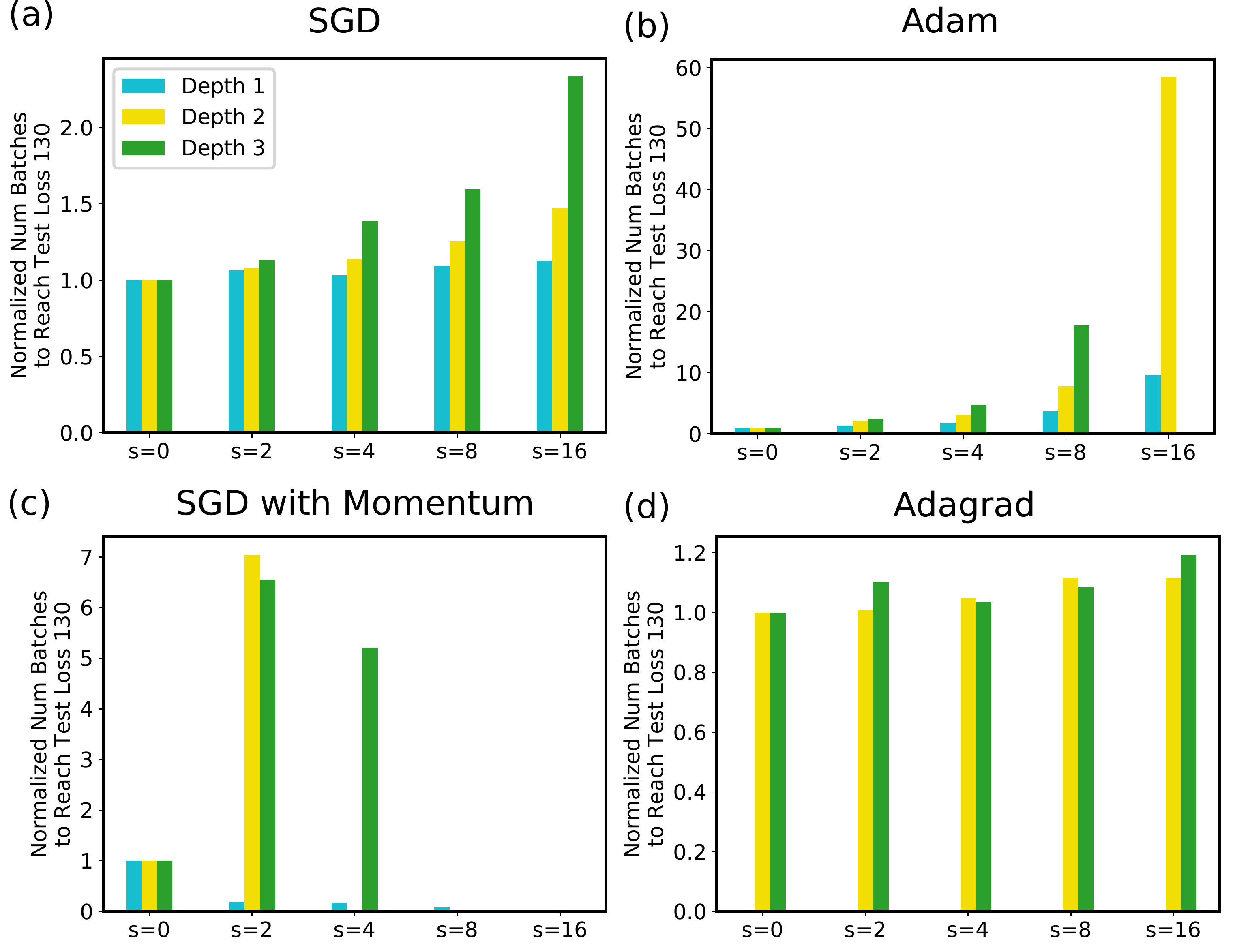}
\caption{The number of batches to reach test loss 130 by Variational Autoencoders (VAEs) on 1 worker, under staleness 0 to 16. We consider VAEs with depth 1, 2, and 3 (the number of layers in the encoder and the decoder networks). The numbers of batches are normalized by $s=0$ for each VAE depth, respectively. Configurations that do not converge to the desired test loss are omitted, such as Adam optimization for VAE with depth 3 and $s=16$.
%The unnormalized version is provided in the appendix (\cref{fig:hypo4_unnorm_panel}).
}
\label{fig:hypo4_panel}
\end{figure}

\cref{fig:hypo4_panel} shows the number of batches to reach test loss 130 by Variational Autoencoders (VAEs) on 1 worker, under staleness 0 to 16 and 4 SGD variants. We consider VAEs with depth 1, 2, and 3 (the number of layers in the encoder and decoder networks). The number of batches are normalized by $s=0$ for each VAE depth, respectively. See the main text for analyses.

\end{document}